\documentclass[sigconf,pbalance=true]{acmart}
\usepackage{graphicx}
\usepackage{xurl}
\usepackage{placeins}
\setcopyright{none}
\acmDOI{}
\acmISBN{}
\acmConference{}{}{}
\renewcommand\footnotetextcopyrightpermission[1]{}
\newcommand{\Cov}{\operatorname{Cov}}

\title[Beyond Wasserstein Proximity]{Beyond Wasserstein Proximity:\texorpdfstring{\\}{ }Finding Better Transfer Sources}

\author{Gordei Verbii}
\affiliation{\department{School of Mathematical Sciences}\institution{Shanghai Jiao Tong University}\city{Shanghai}\country{China}}
\affiliation{\institution{Stratwise}\country{Singapore}}
\author{Georgii Killinkarov}
\affiliation{\institution{Stratwise}\country{Singapore}}
\author{Vladimir Yafarov}
\affiliation{\institution{Stratwise}\country{Singapore}}

\begin{document}
\begin{abstract}
Source selection often favors similar data, yet marginal proximity need not identify the most useful pretraining source. We motivate a conditional alternative through domain-adaptation theory and a regression example in which broader source variation improves estimation despite increasing Wasserstein distance. We examine five-day cryptocurrency volatility using ten Binance spot assets, six forecaster families, three seeds, and shared temporal cutoffs. Selecting the three farthest rather than nearest Gaussian-Wasserstein sources improves mean persistence-relative skill by 12.52 percentage points for MLP and 20.90 for Transformer, with positive post-hoc paired date-block intervals. Forward pilot selection improves mean skill over random transfer in all six families. Compact recurrent models achieve the lowest absolute mean errors, separating selection utility from forecasting quality. Retrospective five-candidate screening shows additional gains from farthest-first geometry before pilot ranking for MLP and Transformer. We distinguish across-target associations from within-target decisions and evaluate target heterogeneity, training seeds, and temporal-block sensitivity. Findings are exploratory for a fixed historical panel; the screening analysis measures no runtime saving.
\end{abstract}

\maketitle
\pagestyle{empty}\thispagestyle{empty}
\hypersetup{pdfauthor={Gordei Verbii and Georgii Killinkarov and Vladimir Yafarov}}

\section{Introduction}
Newly listed cryptocurrency assets provide short and uneven histories for volatility forecasting. Pretraining on another asset can improve the initialization available to a target forecaster, but choosing a source is a consequential decision: transfer can also preserve irrelevant structure. The practical problem is to identify a useful source for a fixed target, learner, and adaptation budget. Windowed Gaussian-Wasserstein distance offers a tractable summary of differences between time-series distributions~\cite{chen}. It is natural to prefer the closest source. We ask when a more distant source can instead be useful.

Similarity has a sound theoretical rationale. Domain-adaptation bounds relate target risk to source risk, distribution discrepancy, and the error of a hypothesis that must work on both domains~\cite{bendavid2010,redko2017}. Reducing discrepancy tightens these bounds when other terms are controlled; distance alone does not order sources with different predictive relationships or estimation quality. Matching feature marginals can also leave conditional differences unresolved~\cite{zhao2019}. These results explain why similarity may be insufficient, but do not themselves predict a benefit from greater distance.

There is a constructive reason to investigate that benefit. A source with broader informative variation can estimate a shared response more accurately while being farther away in marginal geometry. Finite target adaptation can retain this estimation advantage. Section~\ref{sec:conceptual} demonstrates the tradeoff in a simple regression model. For neural forecasting, a source with different activity or volatility scale might similarly expose transferable temporal structure. This is a mechanism to investigate, not one established by the present experiment. Evidence that initialization-dependent benefits can survive fine-tuning~\cite{yosinski2014}, and that learned data selection can combine similarity with diversity~\cite{ruder}, makes the question credible without implying a universal preference for dissimilarity.

Forecasting across heterogeneous series provides further context. Global models need not require identical marginal distributions~\cite{montero2021global}, and domain-adapted forecasting can share temporal components while retaining domain-specific ones~\cite{jin2022daf}. Their benefits depend on what is shared and on the learning procedure. We therefore investigate the conditional hypothesis that \emph{for a fixed target and finite adaptation budget, more distant Gaussian-Wasserstein sources can yield lower forecast error than nearer sources, with the effect depending on the learner}. A direct comparison of source-selection rules tests this hypothesis more closely than an association across targets of unequal difficulty.

We evaluate ten Binance spot assets, six model families, and three seeds under shared calendar boundaries. The contributions are: (i) relating a statistical mechanism to a learner-specific reversal of nearest-source selection; (ii) separating selection gains from absolute quality using target-only, self-continuation, random-transfer, and HAR-style controls; and (iii) testing geometric screening before predictive validation against a matched random-screening comparator. Screening is retrospective; findings have exploratory, fixed-panel scope.

\section{Related work}
\subsection{Domain adaptation and transferability}
Domain-adaptation theory measures mismatch relative to a hypothesis class or loss~\cite{bendavid2010,mansour2009}; regression-specific work studies regularization and sample reweighting~\cite{cortes2014}. Optimal-transport bounds retain source and joint-risk terms alongside Wasserstein discrepancy~\cite{redko2017}. Their assumptions do not directly certify an overlapping-window Gaussian $W_2$ estimate for neural regression. Conditional shifts can obstruct adaptation despite invariant features~\cite{zhao2019}; negative transfer depends on the procedure and available target information~\cite{wang2019}. Transfer difficulty can also change with the hypothesis class~\cite{hanneke2023}. These strands motivate learner-specific comparisons against target-only training.

Feature-transfer experiments distinguish specialization from optimization effects~\cite{yosinski2014}. Distant tasks can still help relative to random initialization, although that work also reports declining transferability with task distance. It therefore does not establish a farthest-source selection rule. Ruder and Plank~\cite{ruder} learn selection criteria combining similarity and diversity in language tasks. Task-aware transferability measures include LEEP, which scores classifier outputs using target labels~\cite{leep2020}, and LogME, which evaluates feature--label evidence and supports regression~\cite{logme2021}. Our forward pilot selector instead directly measures adapted forecasting error on later target outcomes. It is a validation comparator, not a new transferability score.

\subsection{Dataset geometry and predictive similarity}
For Gaussian distributions, Wasserstein geometry compares means and covariances~\cite{gelbrich}. Chen et al.~\cite{chen} apply it to time-series windows and find useful associations with forecasting loss. MMD supplies a kernel-based alternative for comparing distributions~\cite{mmd}. Marginal feature distances can omit predictive structure. Optimal transport dataset distance incorporates feature--label geometry~\cite{otdd}, while Cross-Learning Score uses bidirectional predictive generalization~\cite{cls}. Our one-direction forward validation is not the full Cross-Learning Score. These approaches provide relevant alternatives, but superiority over them is not tested here. The question is how a common marginal geometry relates to source choices for different learners.

\subsection{Learning across time series}
DeepAR shares a recurrent probabilistic forecaster across related series~\cite{salinas2020deepar}. Montero-Manso and Hyndman~\cite{montero2021global} study global versus local forecasting and complexity advantages of global learning; their results do not guarantee transfer from an arbitrary source with a fixed network. Domain Adaptation Forecaster shares attention components while retaining domain-specific values~\cite{jin2022daf}. We instead study separate single-source initializations followed by fixed target adaptation.

Architecture matters independently of selection. Zeng et al.~\cite{zeng2023linear} show that linear forecasters can outperform the Transformer variants examined on long-horizon benchmarks. This motivates linear models with decomposition and compact recurrent controls without predicting their ranking for five-day volatility.

\subsection{Volatility forecasting and evaluation}
Cryptocurrency volatility has an established econometric literature: Katsiampa~\cite{katsiampa2017bitcoin} compares Bitcoin GARCH specifications, including short- and long-run variance components. Andersen et al.~\cite{andersen2003realized} connect intraday observations with realized-volatility measurement and forecasting; Corsi~\cite{corsi2009har} models persistence across daily, weekly, and monthly horizons. These works motivate temporal baselines and a precise target definition. We forecast a logarithmic five-day RMS of daily returns, with a HAR-style comparator adapted to daily-return predictors. Neither is the intraday realized-volatility quantity of those studies. Patton~\cite{patton2011proxy} shows how proxy noise and loss choice can affect forecast rankings. Our findings concern the specified log proxy, not an invariant ranking for latent variance or trading objectives.

\section{Why a more distant source can help}\label{sec:conceptual}
Consider $X_s\sim\mathcal N(0,a^2)$, $X_t\sim\mathcal N(0,1)$, and a shared response $Y=\beta X+\epsilon$, where $a>0$ and the noise is independent, mean-zero, with variance $\sigma^2>0$. Fit a through-origin source least-squares slope $\widehat\beta_s$ using $n>2$ independent observations. Initialize $\beta_0=\widehat\beta_s$ and apply $K$ population-gradient steps on target half-squared loss, with $0<\eta<1$. For target squared risk $R_t$, we obtain
\begin{align}
W_2(P_s^X,P_t^X)&=|a-1|,\label{eq:toyW}\\
\mathbb E[R_t(\beta_K)-R_t(\beta)]
&=\frac{(1-\eta)^{2K}\sigma^2}{a^2(n-2)}.\label{eq:toyRisk}
\end{align}
Conditional source slope variance is $\sigma^2/\sum_iX_{s,i}^2$, and $\sum_iX_{s,i}^2/a^2\sim\chi_n^2$. The inverse chi-square expectation gives the factor $1/(n-2)$; each target update multiplies slope error by $1-\eta$. Target covariate variance is one, so squared slope error equals excess squared risk.

For $a\ge1$, greater distance accompanies smaller expected excess risk: broader source variation estimates the common slope more accurately, and finite adaptation retains the advantage. For $a<1$, moving farther from the target instead reduces information and increases risk. This illustration establishes a conditional possibility. Shared linear responses, independent samples, and population updates are not assumptions asserted for cryptocurrency markets or neural fine-tuning.

\subsection{The relevant conditioning direction}
Let $W_{st}$ be distance and $L_{st}^{(m)}$ test transfer loss for source $s$, target $t$, and family $m$. Correlating them across targets for a fixed source describes a different question from varying sources for one target. For a uniformly sampled ordered pair,
\begin{equation}\label{eq:covariance}
\begin{split}
\Cov(W,L)={}&\mathbb E_t[\Cov(W,L\mid t)]\\
&+\Cov_t(\mathbb E[W\mid t],\mathbb E[L\mid t]).
\end{split}
\end{equation}
This is the law of total covariance, not a decomposition of average correlations. Between-target difficulty can contribute to pooled associations. We consequently use within-target policy losses as the operational evidence.

\begin{figure*}[t]
\centering\includegraphics[width=\textwidth]{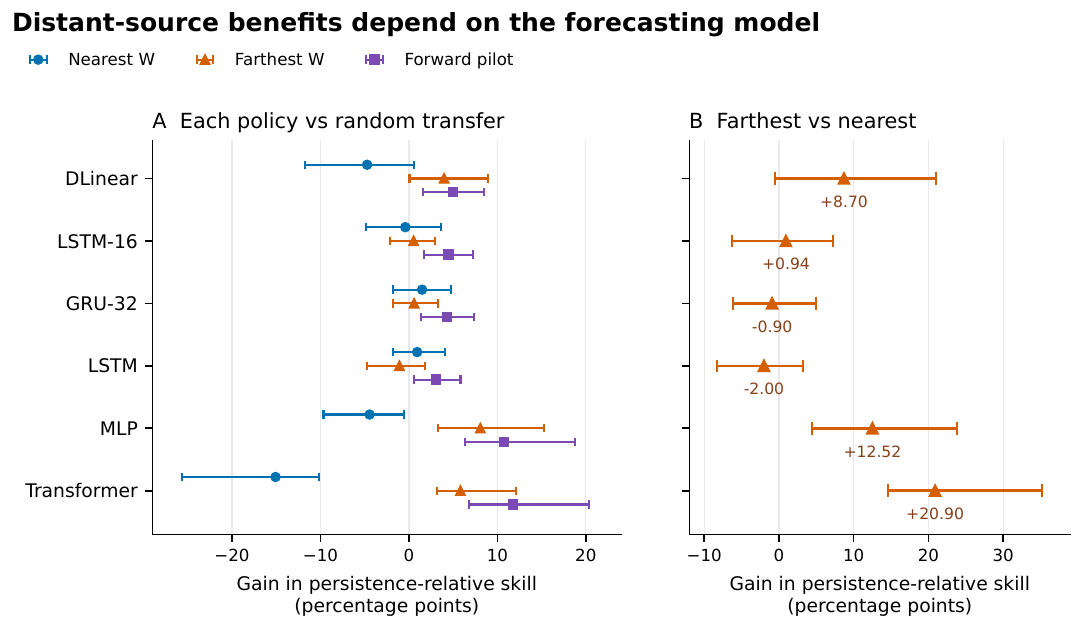}
\caption{Selection effects for three of nine sources. Left: policies versus expected random transfer. Right: direct farthest-versus-nearest contrasts. Positive skill-point effects favor the named policy. Bars are conditional, pointwise 95\% date-block intervals; direct contrasts are post-hoc.}
\Description{Forest plots for six forecasting families. Farthest selection has positive paired intervals versus nearest for MLP and Transformer; pilot versus random intervals are positive for all six.}
\label{fig:selection}
\end{figure*}

\begin{figure*}[t]
\centering\includegraphics[width=\textwidth]{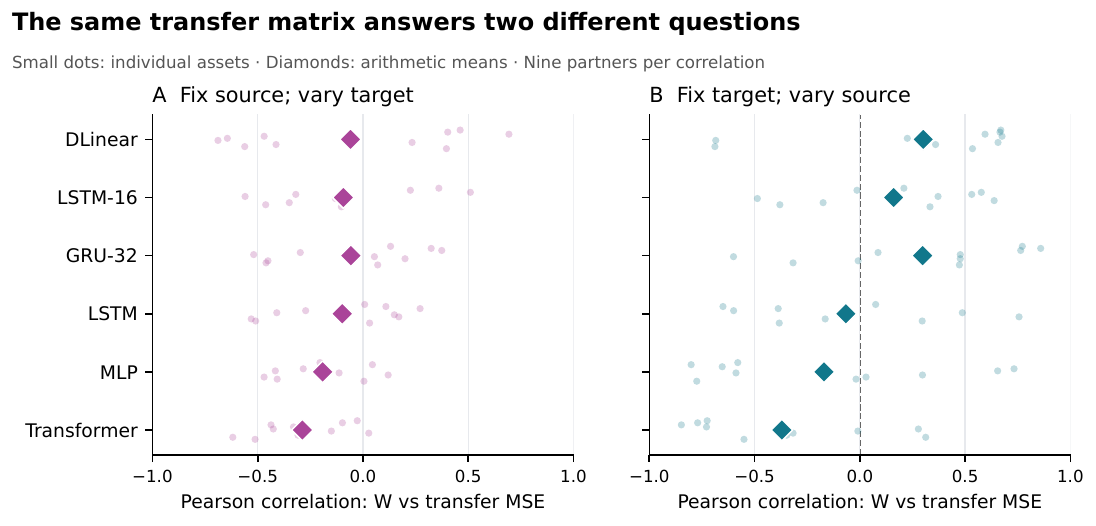}
\caption{Two conditioning directions for the same distance and transfer-loss matrices. Faint dots are correlations across nine partners for each fixed asset; diamonds average over ten assets. These descriptive distributions are not confidence intervals.}
\Description{Per-source correlation means are negative in all families. Per-target means are positive for DLinear and the two compact recurrent models, and negative for LSTM, MLP and Transformer.}
\label{fig:conditioning}
\end{figure*}

\section{Experimental design}
\subsection{Data, target, and information availability}
The frozen panel comprises daily Binance spot USDT bars for BTC, ETH, BNB, XRP, SOL, SEI, JTO, ONDO, SUI, and TIA, from 10 August 2022 through 16 September 2026 where listing histories permit. Public-archive checksums verify the data. We use closes and actual quote-asset volume, with no futures or alternative-provider substitutions. Listing-day bars are retained; rolling warm-up excludes early model origins.

Each timestamp is a UTC bar date. Forecasts are issued after that bar closes. Let $r_{c,u}=\log(p_{c,u}/p_{c,u-1})$. The forward proxy and standardized target are
\begin{align}
v_{c,u}&=\sqrt{\tfrac15\sum_{i=1}^{5}r_{c,u+i}^2},\\
y_{c,u}&=\{\log[\max(10^{-6},v_{c,u})]-\mu_y\}/\sigma_y.
\end{align}
Returns are not clipped. Pooled eligible training labels give $\mu_y=-3.47025894$ and $\sigma_y=0.65780408$. We evaluate squared error on this standardized log scale; common scaling leaves loss ratios unchanged.

The five features are daily log return, absolute log return, trailing 14-day return standard deviation, centered/scaled RSI-14, and log quote volume floored at one before transformation. Predictors use a 60-day lookback. Feature means and standard deviations are fitted to pooled training inputs and frozen. Longer histories contribute more observations to this scaler.

All assets share exclusive boundaries: TRAIN ends 1 October 2025, PILOT ends 1 February 2026, and TEST ends 17 September 2026. Every forward label must end before its stage boundary. Historical feature context may extend into an earlier stage because it is then available. Training uses daily eligible origins. A common five-day phase gives 24 pilot and 45 test origins for every asset; test origins run from 1 February to 9 September 2026, with the final label ending 14 September. There are 1,062 training windows for each of BTC, ETH, BNB, XRP, and SOL; 796 for SUI, 692 for SEI, 615 for TIA, 578 for JTO, and 87 for ONDO.

\subsection{Geometry and training}
For each asset, we flatten 30-day feature windows into $\mathbb R^{150}$ and estimate a Gaussian with Ledoit--Wolf covariance shrinkage. Common preselection feature dates, 3 May 2025 through 31 January 2026, yield 245 overlapping windows. Pilot features already available at selection time enter the score; test features and responses do not. The windows summarize a distribution and are not treated as independent samples for inference. The score is the Gaussian distance~\cite{gelbrich}
\begin{equation}
\begin{split}
W_{st}^2={}&\|\hat\mu_s-\hat\mu_t\|^2+\operatorname{tr}\bigl[\hat\Sigma_s+\hat\Sigma_t\\
&-2(\hat\Sigma_s^{1/2}\hat\Sigma_t\hat\Sigma_s^{1/2})^{1/2}\bigr].
\end{split}
\end{equation}
Distance describes common recent histories, whereas source pretraining uses all eligible training history.

Six scalar-output architectures are evaluated (Table~\ref{tab:models}). Each asset is trained for 40 epochs with Adam at learning rate $10^{-3}$. These fits serve as target-only baselines and source initializations. Every ordered cross-asset transfer then uses ten target epochs at $5\times10^{-4}$ with a fresh optimizer. Self-continuation applies the same ten-epoch adaptation to the target's own 40-epoch fit. It controls extra target training, but is not uninterrupted 50-epoch optimization. Batch size is 64, weight decay $10^{-3}$, gradient-norm clipping 5, dropout zero, and seeds 42, 1234, and 31337. There is no test-driven stopping. The complete design has $6\times3\times(10+10+90)=1,980$ fits. Fixed epochs imply unequal optimizer steps across unequal source histories.

\begin{table}[t]
\caption{Scalar forecasters and trainable parameter counts. DLinear is a scalar-output adaptation with a learned cross-feature readout; recurrent controls are not parameter-matched to it.}
\label{tab:models}
\centering
\begin{tabular}{llr}\toprule
Family & Configuration & Parameters\\\midrule
DLinear & Decomposition-based linear & 616\\
LSTM-16 & One layer, width 16 & 1,489\\
GRU-32 & One layer, width 32 & 3,777\\
LSTM & Two layers, width 64 & 51,521\\
MLP & Two hidden layers, width 128 & 55,169\\
Transformer & Two layers, width 64, 4 heads & 75,009\\
\bottomrule\end{tabular}
\end{table}

Baselines share the test origins: causal five-day RMS persistence, each target's training mean, the pooled training mean, ridge regression on flattened inputs, and HAR-style regression. The latter predicts the log proxy from log absolute return and its trailing five- and 22-day means. Its daily-return inputs differ from the realized-volatility HAR specification~\cite{corsi2009har}.

\subsection{Source choice, metrics, and uncertainty}
For each target and architecture, choose three of nine sources by smallest $W$, largest $W$, or smallest three-seed mean forward pilot transfer MSE. Primary choices are fixed before test evaluation. Losses are averaged over the three separate transfers and seeds; predictions and source data are not combined. Expected random transfer is the exact all-nine mean, also the expected loss of a uniform three-source subset.

For target $c$, let $A_{c,m}^{\pi}$ denote selected mean loss, $R_{c,m}$ random-transfer loss, and $P_c$ persistence MSE. Our primary effect is
\begin{equation}\label{eq:delta}
\Delta_m^{\pi}=\frac1{10}\sum_c\frac{R_{c,m}-A_{c,m}^{\pi}}{P_c}.
\end{equation}
We report $100\Delta$ in percentage points of persistence-relative skill, not percent reduction in reference MSE. Targets are equally weighted after normalization. Macro MSE, target medians, and win counts describe complementary aspects of performance. Paired comparisons with other controls replace $R$ by their losses.

Intervals use 1,000 circular test-date block resamples, four origins per block, and seed 2718. Both arms and all targets use identical sampled dates within a contrast; persistence denominators are recomputed. Primary policy-versus-random summaries use separate draw streams across family--policy combinations; post-hoc paired contrasts share a draw matrix. Percentile 95\% intervals condition on this panel, fitted predictors, averaged seeds, and fixed source choices. They are pointwise and unadjusted for multiplicity, with no retraining, reselection, or asset resampling.

\begin{figure*}[t]
\centering\includegraphics[width=\textwidth]{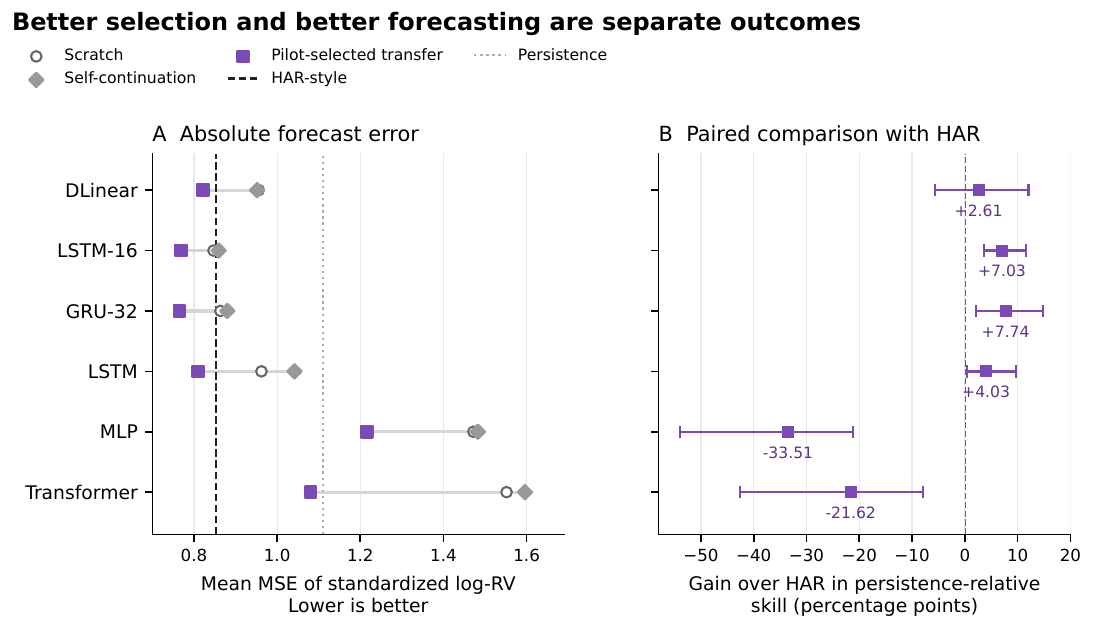}
\caption{Selection utility and forecasting accuracy. Left: macro MSE, where lower is better; scratch uses 40 target epochs and self-continuation adds ten with fresh Adam. Right: post-hoc paired skill gains of pilot-selected transfer over daily-return HAR, with conditional 95\% intervals.}
\Description{Compact recurrent pilot transfers have the lowest mean MSEs. Pilot-selected GRU-32, LSTM-16 and LSTM have positive intervals versus HAR. MLP and Transformer are worse than HAR.}
\label{fig:forecasting}
\end{figure*}

\subsection{Retrospective screening and diagnostics}
We restrict candidates to $q\in\{3,5,7,9\}$, then choose the pilot-best three within each shortlist. Rules retain the nearest $q$, farthest $q$, or a balanced set of $\lceil q/2\rceil$ nearest and $\lfloor q/2\rfloor$ farthest distinct sources. Ties use source-symbol order. The comparator exactly averages over all uniform $q$-source subsets, applying the same pilot selection inside each. There are 84, 126, 36, and one subsets respectively. This comparison separates geometric screening from predictive validation.

All budgets and rules are reported. The $q=5$ contrast is fixed before calculating screening outcomes, but the analysis is post-hoc on an already evaluated panel. Its intervals share four-origin block draws. Candidate counts exclude source pretraining and distance calculation; this retrospective use of all fitted candidates measures no runtime saving.

Further diagnostics compute within-target Spearman correlations of pilot and test source losses and regret to an infeasible test-best-three selector. That selector uses the evaluation outcomes and is an optimistically biased within-candidate reference. Sensitivity checks keep the three-seed pilot choices fixed while evaluating individual seeds and block lengths of two, four, and eight origins.

\begin{figure*}[t]
\centering\includegraphics[width=\textwidth]{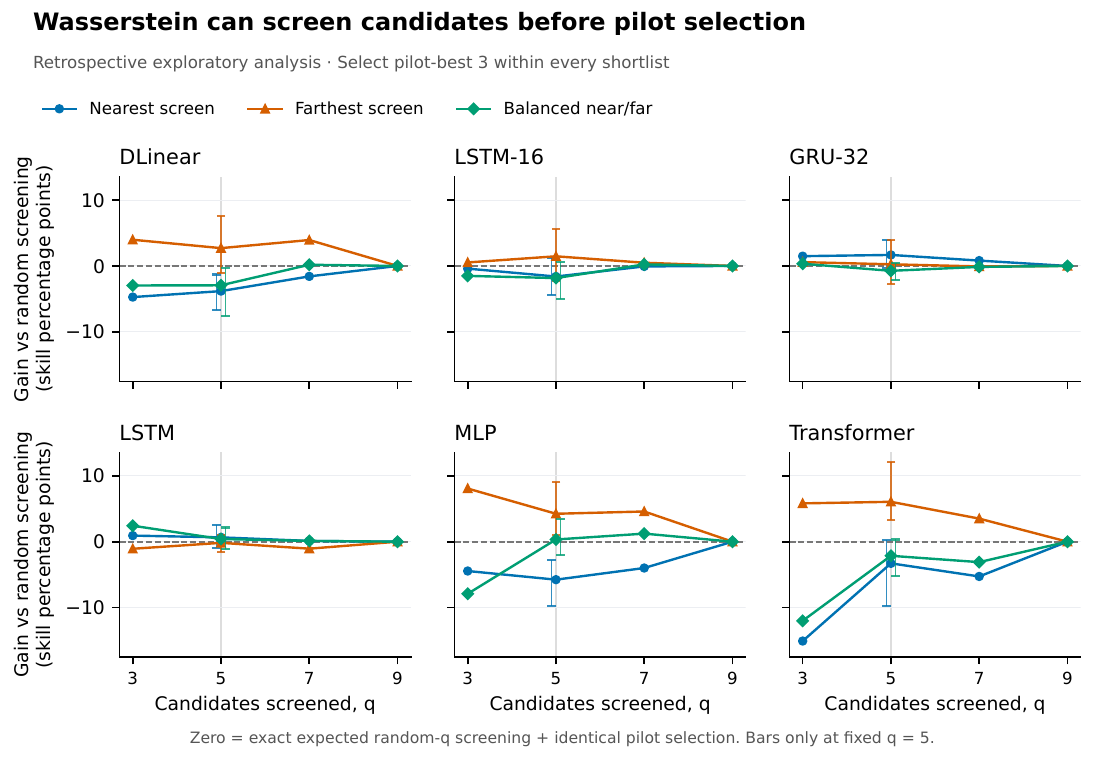}
\caption{Retrospective screening at four fixed candidate budgets. Each shortlist is reduced to its pilot-best three. Zero denotes parity with exact expected random screening of the same size, followed by identical pilot selection. All rules share axes; bars only at $q=5$ show post-hoc conditional 95\% intervals.}
\Description{Six panels compare nearest, farthest and balanced screening with matched random screening. At five candidates, farthest screening has positive intervals for MLP and Transformer. All curves reach zero at nine candidates.}
\label{fig:screening}
\end{figure*}

\section{Results}
\subsection{Operational source-selection effects}
Figure~\ref{fig:selection} compares each selector with expected random transfer and directly contrasts farthest with nearest selection. Farthest-$W$ has positive mean gains for five families, led by MLP at 8.07 points $[3.30,15.27]$. Nearest selection is most adverse in Transformer at $-15.08$ points $[-25.69,-10.15]$. Recurrent families have smaller, mixed geometric effects.

The direct farthest-over-nearest effect is 12.52 points for MLP $[4.47,23.87]$ and 20.90 for Transformer $[14.64,35.21]$. The other four intervals include zero: DLinear 8.70 $[-0.48,21.00]$, LSTM-16 0.94 $[-6.27,7.28]$, GRU-32 $-0.90$ $[-6.16,4.90]$, and LSTM $-2.00$ $[-8.33,3.25]$. These are learner-specific outcomes; different interval signs alone are not a formal between-family interaction test.

Pilot selection improves mean skill over random transfer in every family: 4.97 points for DLinear, 4.49 for LSTM-16, 4.29 for GRU-32, 3.05 for LSTM, 10.73 for MLP, and 11.73 for Transformer. All six conditional intervals are positive. Target effects nevertheless differ. DLinear's nearest rule helps eight targets, but losses on SEI and TIA dominate its negative mean. GRU-32's pilot-over-nearest gain averages 2.81 points, whereas its median is $-2.34$, with three wins, six losses, and one tie. An average gain is not a probability of helping a typical asset.

\subsection{Across-target association and fixed-target choice}
Mean per-source Pearson distance--loss correlations are negative for every family: $-0.059$, $-0.093$, $-0.057$, $-0.098$, $-0.192$, and $-0.288$ in the order of Table~\ref{tab:models}. Yet within-target means are positive for DLinear, LSTM-16, and GRU-32 (Figure~\ref{fig:conditioning}). MLP and Transformer retain negative means in both directions. The spread of asset-specific correlations is substantial. This contrast is consistent with target heterogeneity in Equation~\ref{eq:covariance}, without identifying a causal mechanism. Within-target policy comparisons connect geometry to the source-choice decision.

\subsection{Absolute forecast quality and controls}
Pilot-selected macro MSE is 0.7653 for GRU-32, 0.7689 for LSTM-16, 0.8102 for LSTM, 0.8217 for DLinear, 1.0808 for Transformer, and 1.2155 for MLP. HAR-style MSE is 0.8534 and persistence is 1.1110. Thus strong selection gains for the larger feed-forward and attention models do not make them the best forecasters (Figure~\ref{fig:forecasting}).

Post-hoc pilot-versus-HAR skill effects favor GRU-32 by 7.74 points $[2.06,14.88]$, LSTM-16 by 7.03 $[3.58,11.57]$, and LSTM by 4.03 $[0.47,9.76]$, with nine, seven, and eight target wins respectively. DLinear's 2.61-point mean effect has an interval spanning zero. MLP and Transformer remain worse than HAR in mean error despite benefiting from source selection. No pairwise ranking between the compact recurrent families is established.

Pilot transfer improves mean skill over the corresponding scratch and self-continuation controls in all families, with positive intervals for all twelve contrasts. This statement concerns the specified budgets. DLinear versus scratch averages 11.18 points, but has median $-2.16$ and only four target wins; unequal magnitudes matter. Self-continuation controls additional target epochs but cannot equalize the information in different source histories.

\subsection{Screening candidates before pilot selection}
At $q=5$, farthest screening followed by pilot selection exceeds the matched random-five procedure by 4.24 points for MLP $[1.00,9.06]$ and 6.05 for Transformer $[3.27,12.09]$, with six and eight positive targets. Because random screening also applies pilot-best-three selection, these contrasts measure the incremental contribution of geometry (Figure~\ref{fig:screening}). Other positive five-candidate effects have intervals spanning zero. Nearest-five screening is adverse for DLinear and MLP, and balanced-five screening is adverse for DLinear under these conditional intervals.

All rules coincide with full pilot selection at $q=9$; zero gain means parity of the compared procedures. Five candidates would require $5/9$ of candidate-specific adaptation fits if source parameters were already available, excluding pretraining and distance costs. The analysis reuses the complete candidate sweep and demonstrates no actual compute reduction. It motivates prospective validation of a fixed screening rule, rather than choosing an optimal budget from these test curves.

\subsection{Ranking stability and sensitivity}
Mean within-target pilot--test Spearman correlation ranges from $-0.010$ for GRU-32 to $0.493$ for Transformer. Useful top-three selection can therefore coexist with poor stability of the complete nine-source ranking. Pilot has the smallest mean regret to the infeasible test-best-three selector among the four feasible policies for every family, but residual gaps remain 3.17--11.01 skill points. Mean selected-set overlap is 1.0--1.6 of three; two independent random three-of-nine sets overlap by one source on average. These are descriptive selection gaps, not achievable performance guarantees.

Pilot-versus-random point gains are positive for every seed in every family with the same three-seed pilot choices. Intervals remain positive at block lengths of two, four, and eight origins (10, 20, and 40 calendar days). Farthest-over-nearest intervals remain positive for MLP and Transformer and continue to span zero for the other families. These checks support internal stability under the tested choices, not replication on new assets or regimes.

\section{Discussion and limitations}
The results support learner-specific use of Wasserstein geometry. The illustration explains how informative variation and marginal mismatch can move together; source-history length, response compatibility, optimization, and regularization remain possible contributors in the market experiment. Compact recurrent models combine low error with useful pilot selection, while farthest-first screening is promising for MLP and Transformer.

Ten assets and one test period limit generalization. Temporal information comes from 45 shared origins. Fixed epochs leave unequal data and optimizer budgets; recent-history geometry and full-history pretraining are not disentangled.

Intervals omit retraining and reselection uncertainty. Seed and block sensitivity do not address new-panel generalization. Paired diagnostics and screening are post-hoc, with pointwise intervals unadjusted for multiplicity. Log-proxy MSE does not establish variance calibration, trading profitability, or intraday-volatility accuracy.

Prospective studies should rotate regimes, hold out asset cohorts, match training budgets, and vary adaptation length. Task-aware distances, additional horizons, and tuned GARCH benchmarks would broaden the comparison. Partial-candidate training should measure actual screening costs; multi-source learning should test diversity beyond separate-transfer loss averages.

\section{Conclusion}
Marginal similarity is one component of transfer utility. A simple finite-adaptation example shows how a more distant but informative source can help, motivating a conditional hypothesis for cryptocurrency forecasting. In this fixed panel, farthest-source selection beats nearest selection for MLP and Transformer, forward pilot selection improves mean skill over random transfer in all six families, and compact recurrent models attain the lowest mean errors. Matched-budget screening suggests an incremental role for farthest-first geometry before predictive validation. These findings support learner-specific source selection and motivate prospective tests across assets, regimes, and computational budgets.

\FloatBarrier
\bibliographystyle{ACM-Reference-Format}
\bibliography{references}

@article{bendavid2010,
  author = {Ben-David, Shai and Blitzer, John and Crammer, Koby and Kulesza, Alex and Pereira, Fernando and Vaughan, Jennifer Wortman},
  title = {A theory of learning from different domains},
  journal = {Machine Learning},
  volume = {79}, pages = {151--175}, year = {2010},
  doi = {10.1007/s10994-009-5152-4}
}

@inproceedings{mansour2009,
  author = {Mansour, Yishay and Mohri, Mehryar and Rostamizadeh, Afshin},
  title = {Domain Adaptation: Learning Bounds and Algorithms},
  booktitle = {Proceedings of the 22nd Annual Conference on Learning Theory},
  year = {2009}, publisher = {Omnipress},
  url = {https://www.learningtheory.org/colt2009/papers/003.pdf}
}

@article{cortes2014,
  author = {Cortes, Corinna and Mohri, Mehryar},
  title = {Domain adaptation and sample bias correction theory and algorithm for regression},
  journal = {Theoretical Computer Science}, volume = {519},
  pages = {103--126}, year = {2014}, doi = {10.1016/j.tcs.2013.09.027}
}

@inproceedings{redko2017,
  author = {Redko, Ievgen and Habrard, Amaury and Sebban, Marc},
  title = {Theoretical Analysis of Domain Adaptation with Optimal Transport},
  booktitle = {Machine Learning and Knowledge Discovery in Databases},
  series = {Lecture Notes in Computer Science}, volume = {10535},
  pages = {737--753}, year = {2017}, publisher = {Springer},
  doi = {10.1007/978-3-319-71246-8_45}
}

@inproceedings{zhao2019,
  author = {Zhao, Han and {Tachet des Combes}, Remi and Zhang, Kun and Gordon, Geoffrey},
  title = {On Learning Invariant Representations for Domain Adaptation},
  booktitle = {Proceedings of the 36th International Conference on Machine Learning},
  series = {Proceedings of Machine Learning Research}, volume = {97},
  pages = {7523--7532}, year = {2019}, publisher = {PMLR},
  url = {https://proceedings.mlr.press/v97/zhao19a.html}
}

@inproceedings{wang2019,
  author = {Wang, Zirui and Dai, Zihang and P{\'o}czos, Barnab{\'a}s and Carbonell, Jaime},
  title = {Characterizing and Avoiding Negative Transfer},
  booktitle = {Proceedings of the IEEE/CVF Conference on Computer Vision and Pattern Recognition},
  pages = {11293--11302}, year = {2019},
  url = {https://openaccess.thecvf.com/content_CVPR_2019/html/Wang_Characterizing_and_Avoiding_Negative_Transfer_CVPR_2019_paper.html}
}

@inproceedings{yosinski2014,
  author = {Yosinski, Jason and Clune, Jeff and Bengio, Yoshua and Lipson, Hod},
  title = {How transferable are features in deep neural networks?},
  booktitle = {Advances in Neural Information Processing Systems},
  volume = {27}, pages = {3320--3328}, year = {2014}, publisher = {Curran Associates, Inc.},
  url = {https://proceedings.neurips.cc/paper_files/paper/2014/hash/532a2f85b6977104bc93f8580abbb330-Abstract.html}
}

@inproceedings{ruder,
  author = {Ruder, Sebastian and Plank, Barbara},
  title = {Learning to select data for transfer learning with {Bayesian} Optimization},
  booktitle = {Proceedings of the 2017 Conference on Empirical Methods in Natural Language Processing},
  pages = {372--382}, year = {2017}, publisher = {Association for Computational Linguistics},
  doi = {10.18653/v1/D17-1038}
}

@inproceedings{leep2020,
  author = {Nguyen, Cuong and Hassner, Tal and Seeger, Matthias and Archambeau, Cedric},
  title = {{LEEP}: A New Measure to Evaluate Transferability of Learned Representations},
  booktitle = {Proceedings of the 37th International Conference on Machine Learning},
  series = {Proceedings of Machine Learning Research}, volume = {119},
  pages = {7294--7305}, year = {2020}, publisher = {PMLR},
  url = {https://proceedings.mlr.press/v119/nguyen20b.html}
}

@inproceedings{logme2021,
  author = {You, Kaichao and Liu, Yong and Wang, Jianmin and Long, Mingsheng},
  title = {{LogME}: Practical Assessment of Pre-trained Models for Transfer Learning},
  booktitle = {Proceedings of the 38th International Conference on Machine Learning},
  series = {Proceedings of Machine Learning Research}, volume = {139},
  pages = {12133--12143}, year = {2021}, publisher = {PMLR},
  url = {https://proceedings.mlr.press/v139/you21b.html}
}

@inproceedings{hanneke2023,
  author = {Hanneke, Steve and Kpotufe, Samory and Mahdaviyeh, Yasaman},
  title = {Limits of Model Selection under Transfer Learning},
  booktitle = {Proceedings of the Thirty Sixth Conference on Learning Theory},
  series = {Proceedings of Machine Learning Research}, volume = {195},
  pages = {5781--5812}, year = {2023}, publisher = {PMLR},
  url = {https://proceedings.mlr.press/v195/hanneke23c.html}
}

@inproceedings{otdd,
  author = {Alvarez-Melis, David and Fusi, Nicolo},
  title = {Geometric Dataset Distances via Optimal Transport},
  booktitle = {Advances in Neural Information Processing Systems},
  volume = {33}, pages = {21428--21439}, year = {2020}, publisher = {Curran Associates, Inc.},
  url = {https://proceedings.neurips.cc/paper/2020/hash/f52a7b2610fb4d3f74b4106fb80b233d-Abstract.html}
}

@article{salinas2020deepar,
  author = {Salinas, David and Flunkert, Valentin and Gasthaus, Jan and Januschowski, Tim},
  title = {{DeepAR}: Probabilistic forecasting with autoregressive recurrent networks},
  journal = {International Journal of Forecasting},
  volume = {36},
  number = {3},
  pages = {1181--1191},
  year = {2020},
  doi = {10.1016/j.ijforecast.2019.07.001}
}

@article{montero2021global,
  author = {Montero-Manso, Pablo and Hyndman, Rob J.},
  title = {Principles and algorithms for forecasting groups of time series: Locality and globality},
  journal = {International Journal of Forecasting},
  volume = {37},
  number = {4},
  pages = {1632--1653},
  year = {2021},
  doi = {10.1016/j.ijforecast.2021.03.004}
}

@inproceedings{jin2022daf,
  author = {Jin, Xiaoyong and Park, Youngsuk and Maddix, Danielle and Wang, Hao and Wang, Yuyang},
  title = {Domain Adaptation for Time Series Forecasting via Attention Sharing},
  booktitle = {Proceedings of the 39th International Conference on Machine Learning},
  series = {Proceedings of Machine Learning Research},
  volume = {162},
  pages = {10280--10297},
  publisher = {PMLR},
  year = {2022},
  url = {https://proceedings.mlr.press/v162/jin22d.html}
}

@article{zeng2023linear,
  author = {Zeng, Ailing and Chen, Muxi and Zhang, Lei and Xu, Qiang},
  title = {Are Transformers Effective for Time Series Forecasting?},
  journal = {Proceedings of the AAAI Conference on Artificial Intelligence},
  volume = {37},
  number = {9},
  pages = {11121--11128},
  year = {2023},
  doi = {10.1609/aaai.v37i9.26317}
}

@article{katsiampa2017bitcoin,
  author = {Katsiampa, Paraskevi},
  title = {Volatility estimation for {Bitcoin}: A comparison of {GARCH} models},
  journal = {Economics Letters},
  volume = {158},
  pages = {3--6},
  year = {2017},
  doi = {10.1016/j.econlet.2017.06.023}
}

@article{andersen2003realized,
  author = {Andersen, Torben G. and Bollerslev, Tim and Diebold, Francis X. and Labys, Paul},
  title = {Modeling and Forecasting Realized Volatility},
  journal = {Econometrica},
  volume = {71},
  number = {2},
  pages = {579--625},
  year = {2003},
  doi = {10.1111/1468-0262.00418}
}

@article{corsi2009har,
  author = {Corsi, Fulvio},
  title = {A Simple Approximate Long-Memory Model of Realized Volatility},
  journal = {Journal of Financial Econometrics},
  volume = {7},
  number = {2},
  pages = {174--196},
  year = {2009},
  doi = {10.1093/jjfinec/nbp001}
}

@article{patton2011proxy,
  author = {Patton, Andrew J.},
  title = {Volatility forecast comparison using imperfect volatility proxies},
  journal = {Journal of Econometrics},
  volume = {160},
  number = {1},
  pages = {246--256},
  year = {2011},
  doi = {10.1016/j.jeconom.2010.03.034}
}

@misc{chen,
  author = {Chen, Hongjie and Mehra, Akshay and Kimball, Josh and Rossi, Ryan A.},
  title = {Measuring Time-Series Dataset Similarity using {Wasserstein} Distance},
  year = {2025}, eprint = {2507.22189}, archivePrefix = {arXiv},
  doi = {10.48550/arXiv.2507.22189}, note = {Version 1}
}

@article{mmd,
  author = {Gretton, Arthur and Borgwardt, Karsten M. and Rasch, Malte J. and Sch{\"o}lkopf, Bernhard and Smola, Alexander},
  title = {A Kernel Two-Sample Test},
  journal = {Journal of Machine Learning Research},
  volume = {13}, number = {25}, pages = {723--773}, year = {2012},
  url = {https://jmlr.org/papers/v13/gretton12a.html}
}

@misc{cls,
  author = {Sun, Shudong and Zhang, Hao Helen and Watkins, Joseph C.},
  title = {Quantifying Data Similarity Using Cross Learning},
  year = {2026}, eprint = {2510.10866}, archivePrefix = {arXiv},
  doi = {10.48550/arXiv.2510.10866}, note = {Version 3}
}

@article{gelbrich,
  author = {Gelbrich, Matthias},
  title = {On a Formula for the {$L^2$} {Wasserstein} Metric between Measures on {Euclidean} and {Hilbert} Spaces},
  journal = {Mathematische Nachrichten},
  volume = {147}, number = {1}, pages = {185--203}, year = {1990},
  doi = {10.1002/mana.19901470121}
}
\end{document}